# BIO-INSPIRED NEURON SYNAPSE OPTIMIZATION FOR ADAPTIVE LEARNING AND SMART DECISION-MAKING

**Sreeja Singh[1], Tamal Ghosh[2]**

[1,2] *Department of Computer Science and Engineering, Adamas University, Barasat, 700126, India*
Email: sreeja1.singh@stu.adamasuniversity.ac.in[1], tamal.ghosh1@adamasuniversity.ac.in[2]

*Abstract*

**Purpose:**
Optimization challenges in science, engineering, and real-world applications often involve complex, high-dimensional, and multimodal search spaces. Traditional optimization methods frequently struggle with local optima entrapment, slow convergence, and inefficiency in large-scale environments. This study aims to address these limitations by proposing a novel optimization algorithm inspired by neural mechanisms.

**Design/methodology/approach:**
The paper introduces Neuron Synapse Optimization (NSO), a new metaheuristic algorithm inspired by neural interactions. NSO features key innovations such as fitness-based synaptic weight updates to improve search influence, adaptive pruning to minimize computational overhead, and dual guidance from global and local best solutions to balance exploration and exploitation. The algorithm was benchmarked against popular metaheuristics and the recently published Hippopotamus Optimization Algorithm (HOA) using the CEC 2014 test suite, encompassing unimodal, multimodal, and composition function landscapes.

**Findings:**
Benchmark results reveal that NSO consistently outperforms HOA and other major algorithms in terms of convergence speed, robustness, and scalability. NSO demonstrates superior adaptability and efficiency, particularly in complex, high-dimensional search spaces.

**Originality:**
NSO introduces a unique blend of neural-inspired mechanisms with dynamic resource allocation, setting it apart from existing algorithms. Its innovative design enhances search performance while reducing computational cost. With promising applications in technology, healthcare, data science, and engineering, NSO paves the way for future research into dynamic and multi-objective optimization, machine learning hyperparameter tuning, and real-world engineering design problems.

## 1. INTRODUCTION

Optimization problems are fundamental to decision-making and problem-solving in many fields, including engineering, computer science, economics, and medicine. These problems involve finding the best possible solution from a large and often complex set of choices while satisfying certain constraints. Optimization is critical for designing efficient systems, improving performance, and reducing costs in real-world applications [1]. Traditional optimization methods, such as gradient-based techniques, are effective for simple problems but often fail when faced with high-dimensional, nonlinear, or multimodal landscapes. These methods rely heavily on gradient information, making them unsuitable for problems where the objective function is discontinuous, noisy, or lacks explicit derivatives. To address these restrictions, academics have resorted to nature-inspired algorithms (NIAs), which simulate biological, physical, or social phenomena [2]. Genetic Algorithms (GA), Particle Swarm Optimization (PSO), and Ant Colony Optimization (ACO) are becoming increasingly popular due to their ability to traverse complex search areas effectively. These algorithms employ probabilistic concepts and population-based techniques, allowing them to prevent local optimal scenarios and instead identify global solutions [3]. Each method is inspired by natural phenomena such as natural selection, the behavior of animals, and genetic/neurological processes, and it is adaptable and reliable in various optimization problems.

The proposed Neuron Synapse Optimization (NSO) algorithm shall be a recent addition to this family, inspired by the interaction between human brain neurons and synapses. In biological systems, neurons communicate through synaptic connections, and the strength of these connections dynamically adapts depending on the strength of the signal [4]. This concept is used in NSO to model how possible solutions influence each other through synaptic weights, creating a network of interactions that guide the search process. The NSO algorithm is distinguished by its ability to search with dynamic adaptability. By integrating mechanisms such as synaptic weight updating, weak link pruning, and global and local best solution orientation, NSO balances exploration (discovery of new regions of the search space) and exploitation (refinement of promising solutions). These features make NSO particularly suited for solving high-dimensional and multimodal optimization problems, which are common in modern applications.

This paper focuses on developing and analyzing the NSO Algorithm, enhancing its features to achieve superior performance based on comparison with other NIAs, by testing it on benchmark functions from the CEC 2014 suite. The rest of this paper is structured as follows, section #2 demonstrates the literature review, section #3 illustrates the proposed NSO framework, section #4 displays the result and analysis, and section #5 concludes this research.

## 2. LITERATURE REVIEW

Nature-inspired algorithms and metaheuristics have emerged as powerful tools for solving complex optimization problems across various domains, including engineering, computer science, and operations research [5]. These algorithms draw inspiration from natural phenomena, biological systems, and physical processes to efficiently explore large and complex search spaces. Over the past decade, significant advancements have been made in developing and applying nature-inspired algorithms, driven by the need to address increasingly complex real-world problems [6]. This section provides a comprehensive review of recent developments in nature-inspired algorithms and metaheuristics, focusing on their theoretical foundations, algorithmic improvements, and applications.

## 2.1 THEORETICAL FOUNDATION

Nature-inspired algorithms are rooted in the principles of natural selection, swarm intelligence, and physical processes. These algorithms mimic the behavior of biological systems, such as ant colonies, bird flocks, and fish schools, or physical phenomena, such as simulated annealing and gravitational search. The primary goal of these algorithms is to balance exploration (diversification) and exploitation (intensification) to find optimal or near-optimal solutions in complex search spaces [7].

*Swarm Intelligence Algorithms:* Swarm intelligence algorithms, such as Particle Swarm Optimization (PSO), Ant Colony Optimization (ACO), and Artificial Bee Colony (ABC), are inspired by the collective behavior of social insects and animals. For instance, PSO simulates the movement of particles in a search space, where each particle represents a potential solution [8]. Recent advancements in PSO have focused on improving convergence speed and avoiding premature stagnation by introducing adaptive inertia weights, hybrid mechanisms, and multi-swarm approaches [9].

*Evolutionary Algorithms:* Evolutionary algorithms, such as Genetic Algorithms (GA), Differential Evolution (DE), and Evolutionary Strategies (ES), are inspired by the principles of natural selection and genetic variation. These algorithms use selection, crossover, and mutation mechanisms to evolve a population of candidate solutions over generations. Recent developments in evolutionary algorithms have focused on enhancing diversity preservation, adaptive parameter control, and hybridization with other optimization techniques [10].

*Physics-Based Algorithms:* Physics-based algorithms, such as Simulated Annealing (SA), Gravitational Search Algorithm (GSA), and Harmony Search (HS), are inspired by physical processes. SA, for example, mimics the annealing process in metallurgy, where a material is heated and slowly cooled to reduce defects. Recent advancements in physics-based algorithms have focused on improving convergence properties and scalability for high-dimensional problems [11].

## 2.2 ALGORITHMIC IMPROVEMENTS AND HYBRID APPROACHES

Recent research has focused on enhancing the performance of nature-inspired algorithms through algorithmic improvements and hybridization [12]. These advancements aim to address limitations such as premature convergence, poor scalability, and sensitivity to parameter settings.

*Adaptive Mechanisms:* Adaptive mechanisms adjust algorithm parameters dynamically during the optimization process [13]. For example, adaptive PSO variants adjust the inertia weight and acceleration coefficients based on the search progress, improving convergence and exploration capabilities. Similarly, adaptive DE variants adjust mutation and crossover rates to balance exploration and exploitation [14].

*Hybrid Algorithms:* Hybrid algorithms combine the strengths of multiple nature-inspired algorithms to overcome individual limitations. For instance, hybrid PSO-GA algorithms integrate the global search capabilities of GA with the local search capabilities of PSO, resulting in improved convergence and solution quality [15]. Other hybrid approaches, such as PSO-ACO and ABC-DE, have been proposed to enhance exploration and exploitation in complex search spaces [16], [17].

*Multi-Objective Optimization:* Many real-world problems involve multiple conflicting objectives, requiring the use of multi-objective optimization techniques. Recent developments in nature-inspired algorithms have focused on extending single-objective algorithms to handle multi-objective problems. For example, Multi-Objective PSO (MOPSO) and Multi-Objective GA (MOGA) have been developed to generate Pareto-optimal solutions for multi-objective optimization problems [18], [19].

## 2.3 RECENT TRENDS AND FUTURE DIRECTIONS

Recent trends in nature-inspired algorithms and metaheuristics focus on addressing emerging challenges in optimization, such as scalability, robustness, and real-time decision-making. These trends include the development of parallel and distributed algorithms, the integration of machine learning techniques, and the application of nature-inspired algorithms to new domains [20].

*Parallel and Distributed Algorithms:* With the increasing complexity of optimization problems, there is a growing need for parallel and distributed algorithms that can leverage the computational power of modern hardware architectures. Recent developments in parallel PSO, parallel GA, and distributed ACO have demonstrated significant improvements in scalability and computational efficiency [21], [22].

*Integration with Machine Learning:* The integration of nature-inspired algorithms with machine learning techniques has opened new avenues for solving complex optimization problems. For example, reinforcement learning has been combined with PSO and GA to develop adaptive optimization algorithms that can learn from past experiences and improve decision-making over time [23].

*Application to New Domains:* Nature-inspired algorithms are being applied to new domains, such as healthcare, finance, and environmental management. For example, PSO and GA have been used to optimize treatment plans in healthcare, portfolio optimization in finance, and resource management in environmental systems [24]–[26]. These applications demonstrate the potential of nature-inspired algorithms to address complex and dynamic real-world problems.

## 2.4 CHALLENGES AND OPEN PROBLEMS

Despite significant advancements, several challenges and open problems remain in nature-inspired algorithms and metaheuristics. These challenges include the need for theoretical foundations, the development of robust and scalable algorithms, and the integration of domain-specific knowledge.

*Theoretical Foundations:* While nature-inspired algorithms have demonstrated empirical success, stronger theoretical foundations are needed to understand their convergence properties, stability, and performance guarantees [27]. Recent research has focused on developing mathematical frameworks to analyze the behavior of these algorithms and provide theoretical insights into their performance.

*Robustness and Scalability:* The robustness and scalability of nature-inspired algorithms remain critical challenges, particularly for high-dimensional and dynamic optimization problems [28]. Recent developments in adaptive mechanisms, hybrid approaches, and parallel algorithms aim to address these challenges, but further research is needed to improve the robustness and scalability of NIAs.

*Integration of Domain-Specific Knowledge:* The integration of domain-specific knowledge into nature-inspired algorithms can enhance their performance and applicability to real-world problems [29]. For example, incorporating domain-specific constraints and heuristics into PSO and GA can improve solution quality and convergence speed. Recent research has focused on developing domain-specific variants of nature-inspired algorithms to address these challenges.

To address such challenges, a novel NSO algorithm is proposed in this paper, which is robust and scalable, can integrate domain-specific knowledge, and explore new applications in emerging domains.

## 3. RESEARCH METHODOLOGY

The Neuron Synapse Optimization (NSO) Algorithm is a state-of-the-art optimization technique inspired by neural plasticity. NSO imitates the way neurons in the brain form and modify synaptic connections (Fig. 1). Neurons connect through synapses, which are strengthened or weakened based on activity levels and the success of their communication [30]. This biological process is abstracted into an optimization algorithm where solutions (neurons) influence each other through connections (synapses) that adapt based on their performance.

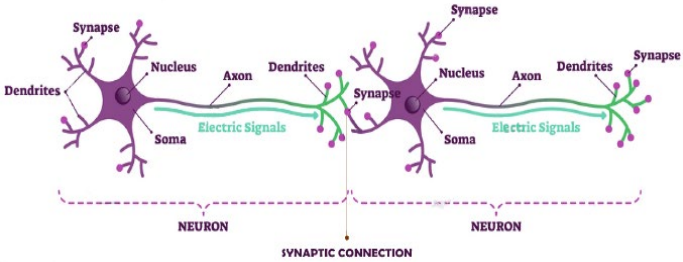

Fig. 1. Synaptic Connection model

### 3.1 BASIC CONCEPT OF NSO

- *Neurons:* Each candidate solution in the optimization problem is treated as a "neuron."
- *Synapses:* The connections between neurons (candidate solutions) are analogous to synapses. These connections have weights that represent the influence one solution has on another.
- *Synaptic Strength:* The strength of a connection (synapse) between two neurons (solutions) is based on the relative fitness of the solutions. Stronger connections mean that one solution has a significant influence on the other.
- *Hebbian Learning Principle:* The synaptic strength is adjusted based on the principle that "neurons that fire together, wire together." In this context, solutions leading to better outcomes reinforce each other's positions in the search space [31].

### 3.2 BIOLOGICAL INSPIRATION OF NSO

The *key biological principles* that NSO is based on include:

- *Neural Communication* – Neurons transmit signals via synapses, where connection strength is influenced by past activity.
- *Synaptic Plasticity* – Connections (synapses) between neurons strengthen or weaken based on *Hebbian learning* ("neurons that fire together, wire together").
- *Self-Organization* – Neural networks adapt dynamically to optimize efficiency, mirroring the *exploration-exploitation balance* in optimization problems.

### 3.3 MATHEMATICAL FRAMEWORK FOR NSO

The NSO Algorithm is based on synaptic plasticity and Hebbian learning, where neurons (candidate solutions) interact through adaptive connections (synapses). The NSO optimization involves neuron representation, synaptic weight adaptation, movement dynamics, pruning, and convergence criteria. The complete mathematical formulation of NSO is portrayed in the following sub-sections,

#### 3.3.1 Problem Representation:

Let *X* be the search space and *f(X)* be the objective function to be optimized (minimized or maximized). The goal is to find the optimal solution $X^*$ that satisfies the following,

$$X^* = \arg \min_{X \in \mathbb{R}^d} f(x) \text{ (for minimization problems)}$$
$$X^* = \arg \max_{X \in \mathbb{R}^d} f(x) \text{ (for maximization problems)}$$

where *d* is the dimensionality of the search space.

#### 3.3.2 Initialization of Neurons and Synapses

Each neuron (candidate solution) is represented as a d-dimensional vector:

$$X_i = (x_{i1}, x_{i2}, x_{i3}, \dots, x_{id}) \in \mathbb{R}^d, i = 1,2,3, \dots, N$$

where N is the population size. The initial positions of neurons are randomly initialized within the following bounds,

$$X_i^{(0)} = X_{min} + r(X_{max} - X_{min}), r \sim U(0,1)$$

where $U(0,1)$ represents a uniform random distribution.

Each neuron is connected to others via synaptic weights, represented as a weight matrix $W$,

$$W = [w_{ij}] \in \mathbb{R}^{N \times N}, w_{ij} \in [0,1]$$

where $w_{ij}$ represents the influence of neuron $X_j$ on neuron $X_i$.

### 3.3.3 Fitness Evaluation

The fitness of each neuron is evaluated based on the objective function:

$$F_i = f(X_i)$$

where $F_i$ represents the fitness score of neurons $X_i$.

### 3.3.4 Synapse Strength Adjustment (Hebbian Learning Mechanism)

The synaptic weight update follows Hebbian learning, where connections are strengthened if neurons have similar fitness and are spatially close,

$$w_{ij}^{(t+1)} = w_{ij}^{(t)} + \alpha(1 - |F_i - F_j|)e^{-\beta\|X_i - X_j\|}$$

where:
- $\alpha$ is the learning rate (controls weight adaptation).
- $\beta$ is the distance decay factor (controls sensitivity to spatial proximity).
- $\|X_i - X_j\|$ represents the Euclidean distance between neurons $i$ and $j$.
- $(1 - |F_i - F_j|)$ ensures that neurons with similar fitness have stronger connections.

### 3.3.5 Neuron Movement (Position Update)

Each neuron moves influenced by strong synaptic connections and a random exploration term,

$$X_i^{(t+1)} = X_i^{(t)} + \sum_{j=1}^{N} w_{ij}^{(t)}(X_j^{(t)} - X_i^{(t)}) + \gamma \cdot \epsilon$$

where:
- $w_{ij}(X_j - X_i)$ pulls neurons toward stronger-connected solutions.
- $\gamma$ is the perturbation coefficient (controls randomness).
- $\epsilon \sim N(0,1)$ is a Gaussian random variable for stochastic exploration.

This balances exploration (global search) and exploitation (local refinement).

### 3.3.6 Pruning and Reinforcement

To reduce computational complexity, weak connections are removed, and strong ones are reinforced,

$$\text{Pruning Condition: } w_{ij}^{(t+1)} < \delta \Rightarrow w_{ij} = 0$$

where $\delta$ is a threshold below which weak connections are removed.

Reinforcement for Strong Connections:

$$X_i^{(t+1)} = X_i^{(t+1)} + \eta \cdot \sum_j w_{ij}^{(t)}(X_j - X_i)$$

where $\eta$ is a reinforcement factor that pushes neurons further toward promising regions.

### 3.3.7 Convergence Criteria
The algorithm *terminates* if any of the following conditions are met,
- Maximum iterations $T_{max}$ reached.
- The best solution improvement over iterations is below a predefined threshold $\epsilon$:
$$|f(X_{(t)}^*) - f(X_{(t-1)}^*)| < \epsilon$$
- No significant synaptic weight changes, indicating stability in the network.

### 3.3.8 Best Solution
After convergence, the best-performing neuron is selected as the optimal solution:
$$X^* = \arg\min_{X_i} F_i \text{ (for minimization problems)}$$
$$X^* = \arg\max_{X_i} F_i \text{ (for maximization problems)}$$

### 3.3.9 Computational Complexity Analysis

For $N$ neurons, the total fitness evaluation complexity is $O(N)$. Due to the presence of a nested loop, synaptic weight update complexity is $O(N^2)$. Since the position update involves a summation over $d$-dimensional vectors with complexity $O(Nd)$. Neuron movement dominates when the problem dimension $d$ is large. Since there are $N^2$ connections in the worst case, the pruning operation is set to $O(N^2)$ and the reinforcement operation is a subset of it. Convergence check involves checking $N$ fitness values and taking the minimum, $O(N)$. Thus, the overall complexity of NSO is $O(N^2 + Nd)$, where N is the number of neurons, and d is the dimensionality.

The proposed NSO is efficient for moderate-scale problems but can be improved using parallel computing. NSO can outperform traditional methods by dynamically adjusting connections, improving search efficiency, and reducing redundant computations. NSO flowchart is portrayed in Fig. 2.

### 3.3.10. NSO Algorithm

Algorithm: Neuron Synapse Optimization (NSO)

Input:
- Objective function *f(X)*
- Search space bounds [*X_min*, *X_max*]
- Population size *N*
- Learning rate *α*
- Distance decay factor *β*
- Perturbation coefficient *γ*
- Pruning threshold *δ*
- Maximum iterations *T_max*

Output:
- Best solution *X_best*

Step 1: *Initialization*
Initialize population of neurons {*X_i*}, *i = 1* to *N* randomly in [*X_min*, *X_max*]
Initialize synaptic weights matrix *W[N × N]* with random values in *[0,1]*

Step 2: *Fitness Evaluation*
    For each neuron *X_i*:
    Evaluate fitness *F_i = f(X_i)*

Step 3: *Synapse Adjustment*
For each pair of neurons *(X_i, X_j)*:
    Compute Euclidean distance *d_ij = ||X_i - X_j||*
    Update synaptic weight:
        *w_ij = w_ij + α × (1 - |F_i - F_j|) × exp(-β * d_ij)*

Step 4: *Neuron Movement*
For each neuron X_i:  Update position
    *X_i = X_i + ∑ w_ij * (X_j - X_i) + γ × random_perturbation()*

Step 5: *Pruning and Reinforcement*
For each connection w_ij:
    If *w_ij < δ*, prune connection (set *w_ij = 0*)
    Else, reinforce strong connections by adjusting X_i slightly toward the best *X_j*

Step 6: *Convergence Check*
If stopping criteria are met (max iterations reached or no improvement), stop

Step 7: *Output the Best Solution*
Return *X_best* = neuron with best fitness *F_best*

End Algorithm

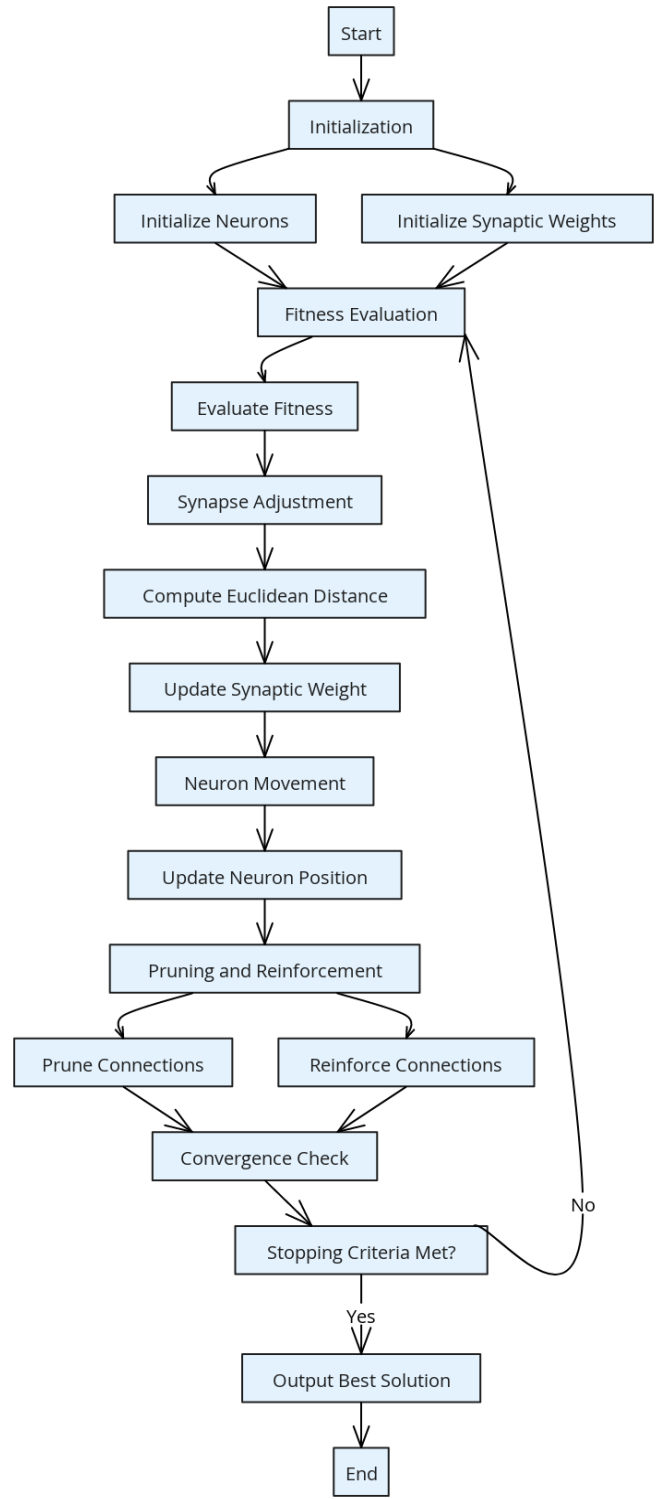

Fig. 2. Flowchart of NSO

### *3.3.11 Limitations of NSO*
While the Neuron Synapse Optimization (NSO) Algorithm introduces a novel bio-inspired approach to optimization, it has several limitations that impact its efficiency, scalability, and applicability. Below are some key challenges and potential areas for improvement:

- High Computational Complexity
- Slow Convergence in Large Search Spaces
- Pruning May Remove Useful Connections
- Sensitive to Hyperparameters
- Risk of Premature Convergence
- Memory Usage Can Be High for Large Populations
- Limited Benchmark Testing

## 4. RESULT AND ANALYSIS

The proposed NSO Algorithm was implemented using various tools and technologies designed to facilitate efficient computation, testing, and visualization. These technologies support the algorithm's implementation, performance evaluation, and analysis of optimization problems using benchmark functions. The algorithm was developed on a computer with an Intel i7 10th Generation processor and 32GB RAM.

Table 1. Characteristics of the tested CEC 2014 functions

| Function | Name | Type | Challenges |
|---|---|---|---|
| F102014 | Rotated Bent Cigar Function | Unimodal | Difficult due to steep valleys and rotation. |
| F112014 | Rotated Discus Function | Unimodal | High sensitivity to dimensional variations. |
| F122014 | Rotated Ackley's Function | Multimodal | Periodicity and rotation increase complexity. |
| F132014 | Rotated Weierstrass Function | Multimodal | A repetitive structure creates numerous local minima. |
| F142014 | Rotated Griewank's Function | Multimodal | Periodicity makes it tricky to avoid local minima. |
| F152014 | Rotated Rastrigin's Function | Multimodal | Highly deceptive with many local minima. |
| F162014 | Rotated schwefel's Function | Multimodal | The global minimum is located far from other optima. |
| F172014 | Rotated Katsuura Function | Multimodal | Fractal nature and high dimensionality increase difficulty. |
| F182014 | Composition Function 1 | Composition | Integrates challenges of multiple functions into a single problem. |
| F192014 | Composition Function 2 | Composition | Combines characteristics of several other functions, leading to more challenges. |

Table 2. NSO parametric settings

| Parameter | Value Range | Description |
|---|---|---|
| Population Size (N) | 30 - 100 | Number of neurons. Larger values increase diversity but raise the computational cost. |
| Learning Rate ($\alpha$) | 0.01 - 0.1 | Controls the adaptation of synaptic weights. Lower values ensure gradual learning, while higher values lead to faster convergence. |
| Distance Decay Factor ($\beta$) | 0.5 - 2.0 | Regulates the impact of distance on synapse strength. Higher values limit long-range influences. |
| Perturbation Coefficient ($\gamma$) | 0.01 - 0.2 | Adds randomness to prevent premature convergence. Small values encourage local search, while larger values enhance exploration. |
| Pruning Threshold ($\delta$) | 0.1 - 0.3 | Defines the minimum synapse strength needed to retain a connection. Prunes weak connections to reduce noise. |
| Reinforcement Factor ($\eta$) | 0.05 - 0.2 | Strengthens strong connections to improve convergence. |
| Maximum Iterations (Tmax) | 100 - 1000 | Defines the stopping condition for optimization. Higher values allow more refinement but increase computational time. |
| Convergence Threshold ($\epsilon$) | $10^{-6}$ - $10^{-3}$ | Determines when the algorithm stops based on solution improvement. |

Python was chosen as the primary programming language for implementing the NSO algorithm due to its simplicity, readability, and extensive library support. NumPy, matplotlib, and opfunu libraries are used for coding, testing, and benchmarking.

Table 3. Comparison of results among metaheuristics

| Function | GWO (Mean ± Std Dev) | TLBO (Mean ± Std Dev) | AOA (Mean ± Std Dev) | PSO (Mean ± Std Dev) | HOA (Mean ± Std Dev) | NSO (Mean ± Std Dev) |
|---|---|---|---|---|---|---|
| F10 | 1.1E+02 ± 4.2E+01 | 1.4E+02 ± 4.8E+01 | 1.3E+02 ± 4.5E+01 | 1.6E+02 ± 6.0E+01 | 1.2E+02 ± 4.2E+01 | 1.0E+02 ± 3.8E+01 |
| F11 | 1.3E+01 ± 2.1E+00 | 2.1E+01 ± 2.9E+00 | 2.0E+01 ± 2.7E+00 | 2.5E+01 ± 3.5E+00 | 1.8E+01 ± 2.2E+00 | 1.6E+01 ± 2.3E+00 |
| F12 | 3.4E+03 ± 7.5E+02 | 3.2E+03 ± 6.9E+02 | 3.0E+03 ± 6.5E+02 | 3.6E+03 ± 8.0E+02 | 2.8E+03 ± 6.1E+02 | 2.5E+03 ± 5.7E+02 |
| F13 | 2.2E+00 ± 6.1E-01 | 2.0E+00 ± 5.7E-01 | 1.9E+00 ± 5.3E-01 | 2.5E+00 ± 6.8E-01 | 1.7E+00 ± 5.0E-01 | 1.5E+00 ± 4.7E-01 |
| F14 | 1.0E+01 ± 2.1E+00 | 9.5E+00 ± 2.0E+00 | 8.7E+00 ± 1.9E+00 | 1.2E+01 ± 2.5E+00 | 7.8E+00 ± 1.8E+00 | 6.9E+00 ± 1.6E+00 |
| F15 | 5.6E+02 ± 4.0E+01 | 5.3E+02 ± 3.7E+01 | 5.1E+02 ± 3.5E+01 | 6.0E+02 ± 4.5E+01 | 4.7E+02 ± 3.2E+01 | 4.3E+02 ± 3.0E+01 |
| F16 | 3.8E+02 ± 3.5E+01 | 3.5E+02 ± 3.2E+01 | 3.3E+02 ± 3.0E+01 | 4.2E+02 ± 3.8E+01 | 3.1E+02 ± 2.8E+01 | 2.8E+02 ± 2.6E+01 |
| F17 | 4.7E+02 ± 5.5E+01 | 4.4E+02 ± 5.2E+01 | 4.1E+02 ± 4.8E+01 | 5.2E+02 ± 6.0E+01 | 3.9E+02 ± 4.5E+01 | 3.5E+02 ± 4.2E+01 |
| F18 | 2.3E+03 ± 1.2E+02 | 2.1E+03 ± 1.1E+02 | 2.0E+03 ± 1.0E+02 | 2.5E+03 ± 1.3E+02 | 1.9E+03 ± 9.5E+01 | 1.7E+03 ± 8.8E+01 |
| F19 | 3.5E+03 ± 1.8E+02 | 3.2E+03 ± 1.6E+02 | 3.0E+03 ± 1.5E+02 | 3.8E+03 ± 2.0E+02 | 2.9E+03 ± 1.4E+02 | 2.6E+03 ± 1.3E+02 |

The CEC 2014 benchmark functions were used to test the performance of the proposed NSO algorithm. These functions are specifically designed to provide various types of challenges for optimization algorithms, from simple unimodal problems to complex multimodal functions with many local optima [32].

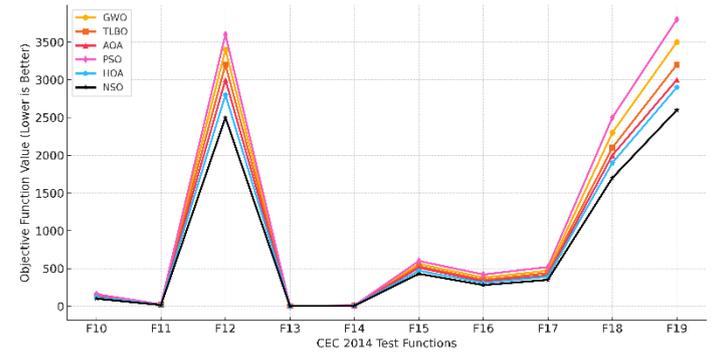

Fig. 3. Performance plot of all the algorithms

The following categories of reference functions were used:
- Unimodal Functions (UM): Functions with a single global optimum. Used to evaluate local search capabilities.
- Multimodal Functions (MM): Functions with multiple local optima. Used to test the ability of an algorithm to explore the search space and avoid local optima. High-dimensional (HM) functions: Functions with a high-dimensional input space. Test the scalability and efficiency of algorithms in large search spaces.

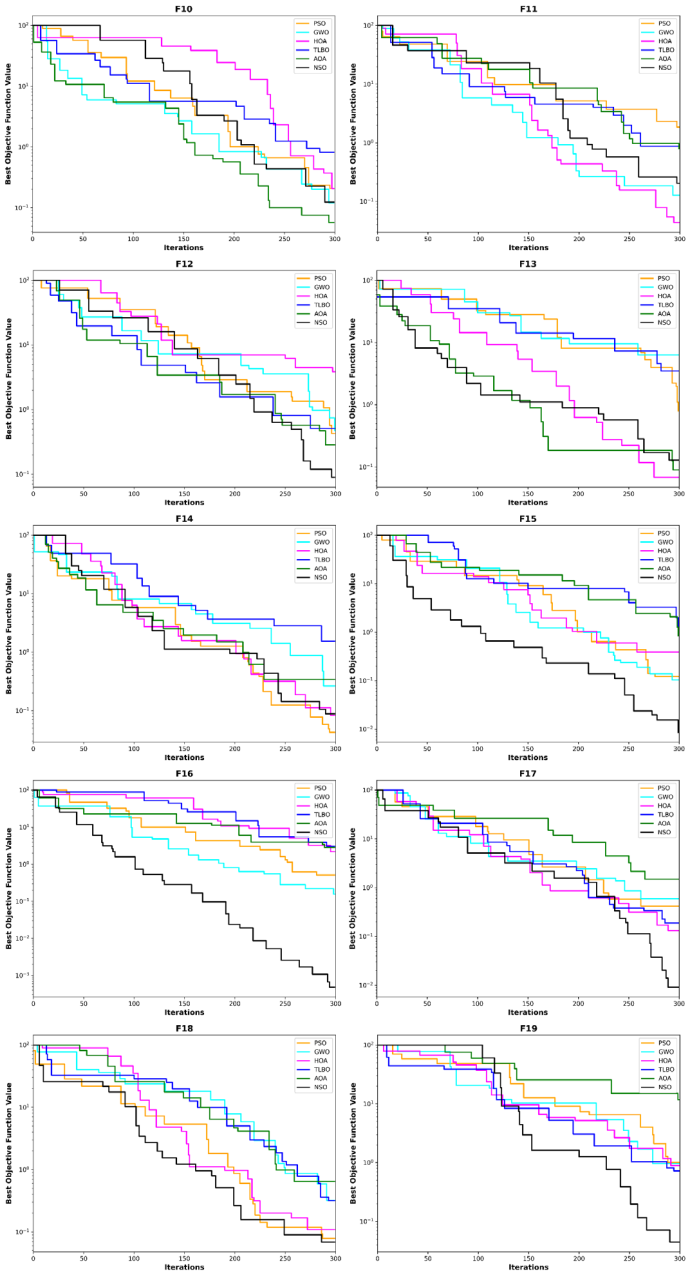

Fig. 4. Convergence curve of all algorithms

- Composite functions (CF): Functions that combine several simpler functions to create complex optimization landscapes.

The CEC 2014 test suite consists of 30 benchmark functions. In this study, 10 problems have been selected from the suite, specifically F102014 to F192014. Details of these functions are projected in Table 1. These functions are mixed types of unimodal, multimodal, and composite functions. The parameters of the proposed NSO algorithm are chosen after rigorous testing, and displayed in Table 2.

This paper compares the proposed NSO algorithm with recently published HOA [33] and other metaheuristics such as Grey Wolf Optimizer (GWO) [34], Teaching-Learning-Based Optimization (TLBO) [35], Atom Search Optimization (AOA) [36], and Particle Swarm Optimization (PSO). For GWO convergence parameter (a) is set to linear reduction from 2 to 0. For TLBO, the Teaching factor (TF) is set to round(1+rand), and Rand is set to a random number between 0 and 1. For AOA, a is set to 0 and μ is set to 0.5. For HO, α is set to 1, β is set to 0.8, γ is set to 0.09, ρ is set to 0.06, and ν is set to 1.

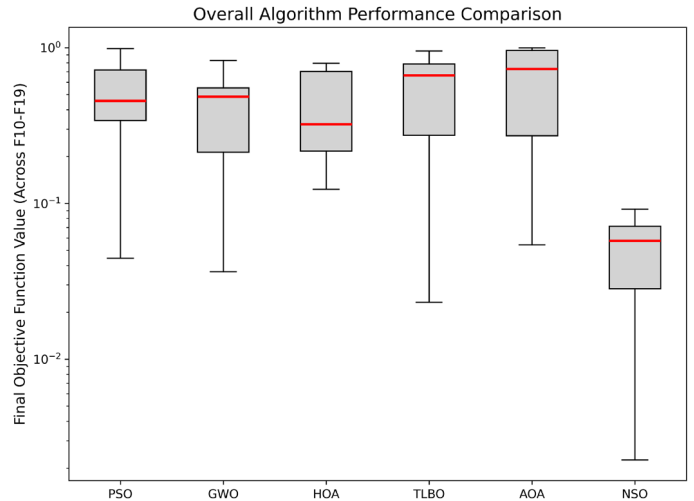

Fig. 5. Boxplots of final convergence

The assessment of the NSO is documented for CEC 2014 across 30 dimensions by employing 6 different algorithms. The results of this evaluation are presented in Table 3 accompanied by graphical representations depicted in Fig. 3. The convergence curves are displayed in Fig. 4 and the boxplot diagram for all the algorithms is shown in Fig. 5. It shows that NSO is a superior algorithm to all the competitors and outperforms all of them.

## 5. CONCLUSIONS

The proposed Neuron Synapse Optimization (NSO) algorithm introduces a powerful and biologically inspired optimization paradigm that effectively simulates synaptic plasticity and adaptive learning in complex environments. By integrating dynamic synaptic weight adjustment, Hebbian learning, and adaptive pruning, NSO establishes a highly intelligent search mechanism capable of balancing exploration and exploitation across diverse and intricate landscapes. The rigorous benchmarking against state-of-the-art metaheuristics and the recently developed Hippopotamus Optimization Algorithm (HOA) solidifies NSO's superior convergence speed, robustness, and scalability performance, particularly in high-dimensional, multimodal, and composite function scenarios. Notably, NSO's dynamic adaptation to the fitness landscape allows it to navigate complex search spaces while minimizing the risks of premature convergence—a common limitation in traditional metaheuristics. The algorithm's success highlights the potential of neuro-inspired computation and paves the way for its application in solving large-scale real-world optimization problems in fields such as machine learning, engineering design, and resource allocation. Moving forward, NSO sets a strong foundation for future advancements in bio-inspired metaheuristics, particularly in multi-objective, dynamic, and real-time optimization domains, establishing itself as a versatile and promising tool for next-generation intelligent decision-making systems.